\documentclass[lettersize,journal]{IEEEtran}
\usepackage{amsmath,amsfonts}
\usepackage{algorithmic}
\usepackage{algorithm}
\usepackage{array}

\usepackage[caption=false,font=footnotesize,labelfont=sf,textfont=sf]{subfig}
\usepackage{textcomp}
\usepackage{stfloats}
\usepackage{url}
\usepackage{verbatim}
\usepackage{graphicx}
\usepackage{cite}

\begin{document}

\title{2T-UNET: A Two-Tower UNet with Depth Clues for Robust Stereo Depth Estimation}
\author{Rohit Choudhary $^{1}$,~Mansi~Sharma $^{1}$ and ~Rithvik Anil $^{2}$
\thanks{$^1$Department of Electrical Engineering,
Indian Institute of Technology Madras, Tamil Nadu 600036, India
(e-mail: ee20s002@smail.iitm.ac.in, mansisharmaiitd@gmail.com, mansisharma@ee.iitm.ac.in)}

\thanks{$^2$Department of Mechanical Engineering, Indian Institute of Technology Madras, Tamil Nadu 600036, India (e-mail: me17b170@smail.iitm.ac.in, rithvik.anil@gmail.com)}
}

\maketitle

\begin{abstract}
Stereo correspondence matching is an essential part of the multi-step stereo depth estimation process. This paper revisits the depth estimation problem, avoiding the explicit stereo matching step using a simple two-tower convolutional neural network. The proposed algorithm is entitled as 2T-UNet. The idea behind 2T-UNet is to replace cost volume construction with twin convolution towers. These towers have an allowance for different weights between them. Additionally, the input for twin encoders in 2T-UNet are different compared to the existing stereo methods. Generally, a stereo network takes a right and left image pair as input to determine the scene geometry. However, in the 2T-UNet model, the right stereo image is taken as one input and the left stereo image along with its monocular depth clue information, is taken as the other input. Depth clues provide complementary suggestions that help enhance the quality of predicted scene geometry. The 2T-UNet surpasses state-of-the-art monocular and stereo depth estimation methods on the challenging Scene flow dataset, both quantitatively and qualitatively. The architecture performs incredibly well on complex natural scenes, highlighting its usefulness for various real-time applications. Pretrained weights and code will be made readily available.
\end{abstract}

\begin{IEEEkeywords}
3D TV, augmented reality, computational photography, convolution neural network, deep learning, stereo depth estimation, virtual reality.
\end{IEEEkeywords}

\maketitle

\section{Introduction}

Depth estimation from RGB images has been a crucial research topic in computer vision and computer graphics. It plays a critical role in the domain of 3D reconstruction, virtual reality, augmented reality \cite{Newcombe2011ICCV, Avinash2019Mobile3DAR}, mapping and localization, image refocusing \cite{Moreno2007RefocussingofImages}, image segmentation \cite{Hazirbas2017}, robotics navigation \cite{Desouza2002MobileRobotNavigation}, autonomous driving \cite{Menze2015AutonomousVehicles}, novel view synthesis \cite{Vineet2020NovelView},  free-viewpoint TV \cite{Sharma2019FTV, Sharma2019FTVKrylov} and 3D displays \cite{Sharma2019MV3DHDR, Joshitha2021GF3DDisplays, Sharma20143Ddisplay}.

In general, stereo depth estimation is a multi-step process where the steps can be classified into four categories: 1) feature generation, 2) feature matching, 3) disparity estimation, and 4) refinement of the disparity. The feature generation step in a standard stereo depth estimation network compute features for both views using a shared-weight CNN. The next step, \textit{i.e.}, feature matching, uses a cost volume to calculate how close the feature maps are at various levels of disparity. Patch correspondence of features extracted from the stereo pair is performed in this step \cite{Survey, OctDPS, PSMNet,MADNet,ATVSNet}. For example, a patch in the left feature map centered around $(i, j)$ is matched with multiple patches centered around $(i, j-d)$ in the right feature map, where $d$ is the disparity level. A similarity rating is computed for each of the pairs at various disparity levels. Current methods consider different metrics for similarity computation, such as $L_1$ norm, $L_2$ norm, cosine similarity, correlation, etc \cite{cosineSim, AnyNet, correlation2, MADNet}. Some methods construct a 4D cost volume by simply appending features at different levels of disparity to the corresponding patch from the left feature map. This cost volume is regressed through several convolution layers to the estimated depth map and refined further to the desired output \cite{PSMNet,DPSNet,GANet,ATVSNet,4D2}.

\begin{figure*}[!t]
\centering
\includegraphics[width=6.8in, height=2.4in]{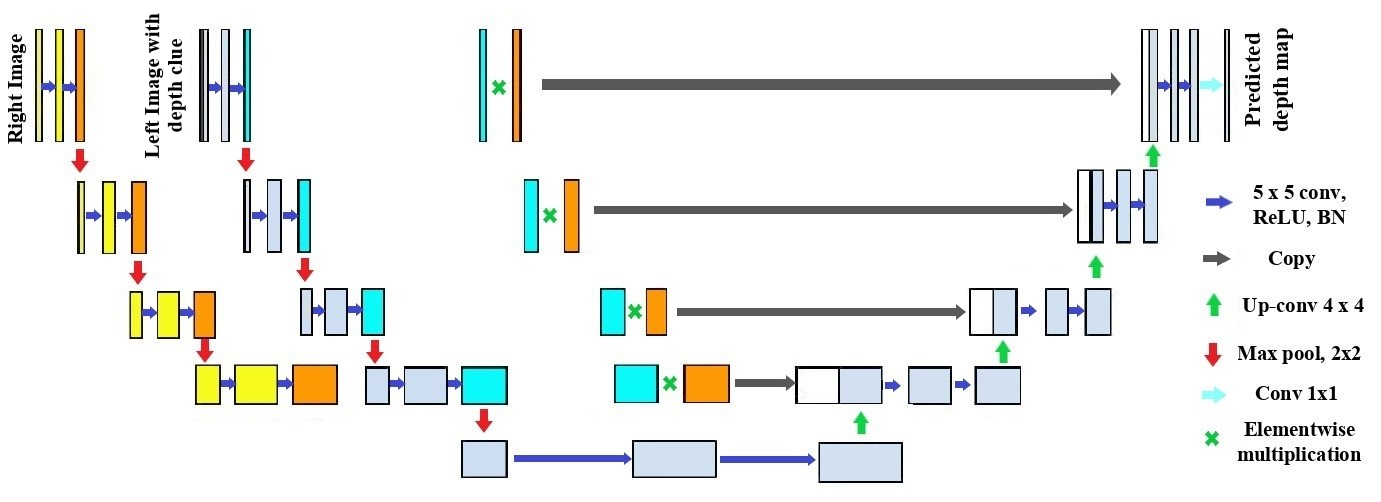}
\caption{Overview of proposed 2T-UNet architecture: The network, inspired by UNet, employs a novel secondary encoder to eliminate the cost volume construction. A novel fusion strategy is introduced, where twin features are fused from the encoder before sending the information to the decoder.}
\label{2T-UNet}
\end{figure*}

The differentiating attribute of our 2T-UNet architecture is that there is no need for explicit cost volume construction. The cost volume moves the features at several incremental levels of disparity to compute the similarity scores  \cite{PSMNet,ATVSNet,4D2,AnyNet}. Alternatively, the twin encoders in 2T-UNet implicitly capture the disparity or depth information by shifting its weights. Removing cost volume by transferring its functions to the encoders offers our network flexibility in deciding the range of disparity variation that should be considered for a given scene. Standard stereo-based depth prediction networks set the highest disparity value based on the training data. However, our method allows the network to infer the highest disparity value on its own \cite{SDE}. This helps us achieve robust depth predictions on tricky scenes with the fine details and clear object boundaries in the depth map. 2T-UNet is also provided with a monocular depth clue, a complementary depth suggestion from a pre-trained network. This idea is inspired by the work of Watson et al. \cite{depthhints}. The depth hints are used in functions in the domain of self-supervised learning, wherein the network predicts disparities of the scene using an RGB stereo pair alone and no depth labels. In the self-supervised depth estimation approach, the stereo pair is reconstructed using predicted disparity. The reconstruction loss with the target views (stereo pair) is critical for better convergence of their network. The depth hints are computed using standard heuristically designed stereo methods \cite{depthhints}.

In the proposed method, the depth clue is obtained using the left view of the stereo pair and passing it through an off-the-shelf monocular depth estimation algorithm. The depth clue is provided to the network by concatenating monocular depth information with the left stereo view. The secondary encoder takes input of the right view alone. The off-the-shelf algorithm is not trained on the same dataset as our network. Our method differs from Watson et al. \cite{depthhints} work in two crucial aspects: 

{\textbf{Application domain}: Our method is on stereo depth estimation domain. However, Watson et al. \cite{depthhints} depth hints work is in the domain of self-supervised monocular depth estimation.} 

{\textbf{Location of application in the network}: Our method provides a coarse depth clue to the network, whereas the depth hint in Watson et al. \cite{depthhints} is used during the loss calculation}.

The 2T-UNet is composed of following four novel ideas: 

{\textbullet \hspace{1 mm} Our architecture uses a twin encoder which is differently weighted \cite{SDE}. This is contrary to the conventional stereo algorithms that generally make use of encoders with shared weights followed by cost volume construction \cite{PSMNet,AnyNet,GANet,DPSNet}.}

{\textbullet \hspace{1 mm} In addition, monocular depth clues are used to aid our network in predicting a high-quality depth map. We computed coarse geometric information of the scene using the algorithm of Ranftl et al. \cite{Midas}. The depth clue is provided as complementary guidance to our network. The depth clues assist the proposed network to better preserve depth discontinuities and retain much better feature definitions in the depth map predictions. We also make sure that depth is learned from features derived from the stereo pair and not directly from the given depth clues. The coarse depth clues used in our proposed 2T-UNet network are depicted in Fig.~\ref{visComp} (d).}
    
{\textbullet \hspace{1 mm} Another distinguishing characteristic of 2T-UNet is that it uses a simple fusion strategy. The features are bitwise multiplied before being sent from encoder to decoder, \textit{i.e.}, each corresponding element of the feature maps are multiplied before concatenating them in the decoder \cite{SDE}. This is a crucial step in the process of replacing cost volume-construction method. 

A standard stereo depth estimation network \cite{PSMNet,DPSNet,GANet,ATVSNet,4D2}, on the other hand, performs the stereo matching step using cost volume. Standard methods achieve this by using an additional parameter $d_{max}$, which depicts the maximum extent to which the stereo feature must be shifted disparity-wise in order to perform stereo matching. This maximum disparity parameter is set explicitly in the model definition based on the variation in the dataset \cite{AnyNet}. To enable our network to match features without explicitly specifying the disparity limit parameter, the feature matching process is re-imagined. 

To make our 2T-UNet network perform the feature matching without explicitly setting the disparity limit parameter, we re-imagine the feature matching step. Instead of shifting the features within the cost volume, we use a functionally identical shift in weights at the encoders to capture the disparity. Thus, to capture disparity information, it is necessary that shift in the weight in either of the encoders influences the other. By multiplying the features elementwise, we are ensuring that the weights are updated in a dependent fashion. Backpropagation through the multiplication operation ensures that weight update happens in tandem between the two encoders\cite{SDE}. 

\begin{table*}[t!]
\small
    \centering
    \caption{Comparison of proposed 2T-UNet scheme with different monocular \textit{(top)} and stereo \textit{(below)} based depth estimation methods on Scene flow dataset. Lower values are better for $abs\_rel$, $sq\_rel$, $log_{10}$ and RMSE. Higher values indicate better quality for $\sigma_1$, $\sigma_2$, $\sigma_3$ and SSIM measures. Best method per metric is highlighted in bold. Second best method per metric is underlined. }
    \begin{tabular}{|c||c|c|c|c|c|c|c|c|}
    \hline
    \textbf{Method}& $abs\_rel$ $\downarrow$ & $sq\_rel$ $\downarrow$ & $log_{10}$ $\downarrow$ & RMSE $\downarrow$ & $\sigma_1$ $\uparrow$ & $\sigma_2$ $\uparrow$ & $\sigma_3$ $\uparrow$ & SSIM $\uparrow$ \\
    
    \hline
    AdaBins \cite{Adabins2021} & 1.222 & 0.414 & 0.283 & 0.264 & 0.240 & 0.411 & 0.569 & 0.644 \\
    CADepth \cite{CADepth2021} & 0.456 & 0.059 & 0.142 & 0.087 & 0.522 & 0.760 & 0.867 & 0.826 \\
    DenseDepth \cite{densedepth} &  1.976 &  0.738 &  0.377 &  0.309 &  0.142 &  0.288 &  0.444 &  0.555  \\ 
    FCRN \cite{FCRN} &  1.143 & 0.288 &  0.280 &  0.219 &  0.208 &  0.394 &  0.599 &  0.658 \\
    Depth Hints \cite{depthhints} &  0.788 &  0.136&  0.208 &  0.136 &  0.338 &  0.635 &  0.763 &  0.769  \\ 
    SerialUNet \cite{SerialNet} &  0.861 &  0.165 &  0.216 &  0.151 &  0.344 &  0.576 &  0.735 &  0.713   \\ 
    MSDN \cite{MSDN} &  0.856  &  0.174 &  0.234 &  0.189 &  0.241 &  0.457 &  0.702 &  0.715  \\ 
    SIDE \cite{SIDE} &  0.958 & 0.218 &  0.239 &  0.169 &  0.325 &  0.540 &  0.707 &  0.726  \\ 
    MiDaS \cite{Midas} &  0.338 &  0.031 &  0.146 &  0.072 &  0.550 &  0.782 &  0.879 &  0.840   \\
    \hline
    OctDPSNet \cite{OctDPS} &  1.552 &  0.498 &  0.422 &  0.319 &  0.139 &  0.275 &  0.407 &  0.509   \\ 
    PSMNet \cite{PSMNet} &  0.317  & 0.022 &  0.226 &  0.062 &  0.501 &  0.665 &  0.773 &  0.781 \\
    DeepROB \cite{DeepPruner_ROB} &  0.245 &  0.016 &  0.155 &  0.049 &  0.622 &  0.766 &  0.840 &  0.816    \\ 
    HSMNet \cite{HSM}&  0.485 &  0.106 &  0.286 &  0.164 &  0.383 &  0.517 &  0.600 &  0.685  \\ 
    STTR \cite{STTR}&  1.016 & 0.342 &  2.341 &  0.410 &  0.003 &  0.005 &  0.008 &  0.018    \\ 
    SDE-DENet \cite{SDE}&  \textbf{0.166} &  \textbf{0.008} & \underline{0.097} &  \textbf{0.031} &  \textbf{0.740} &  \underline{0.858} &  \underline{0.903} &  \underline{0.872}    \\ 
    \hline
    
    \textbf{2T-UNet (ours)}  &  \underline{0.218} & \underline{0.013} &  \textbf{0.084} &  \underline{0.037} &  \underline{0.736} &  \textbf{0.880} &  \textbf{0.935} &  \textbf{0.886}  \\ 
    \hline
    \end{tabular}
\label{Table-comparison}
\end{table*}

Furthermore, in conventional stereo networks, the cost volume covers feature shift at all possible disparity values, and therefore contains large amounts of information. This large information bundle cannot be captured by shifting the weights in a single convolution block. Therefore, in the proposed 2T-UNet, we split workload among a sequence of weight shifts along with the twin encoders with periodic element-wise multiplication operations.}
    
{\textbullet \hspace{1 mm} Convolutional operations are an integral part of a CNN. This operation captures information about a point and its neighbours. For network to capture the disparity between views, we should extend the information extraction to both views. This awareness extension is achieved by multiplying the features from the encoders elementwise, \textit{i.e.}, fusing them into one feature map, before sending them to the decoder. The fusion operation ensures that a feature point is aware of its neighbours and its disparity with the other view \cite{SDE}. In addition, 2T-UNet executes feature fusion at various resolutions, capturing disparity information at different scales. The spread of disparity capture allows depth map to retain minute details and well-preserved object boundaries.}

To summarise, 2T-UNet is inspired by SDE-DualENet (SDE-DENet)\cite{SDE}, but differs significantly in its architecture and certain core ideas. The SDE-DENet \cite{SDE} is based on EfficientNets \cite{EffNet}, while 2T-UNet is built on top of the UNet architecture \cite{UNet}. Both techniques differ significantly in the way in which information from the encoder is added to the decoder. In SDE-DENet, the encoder feature is added elementwise to the decoder feature, resulting in some information loss. In 2T-UNet, however, the encoder features are concatenated to the decoder feature, carrying more information. The distinguishing attribute that sets 2T-UNet apart from SDE-DENet is the usage of monocular depth clues. These clues aid the network converge better and give results with clear object boundaries, while maintaining minute details. There are five main sections that make up the rest of this article. Section \ref{sec:rel_work} covers various depth estimation algorithms. In Section \ref{sec:arch}, the proposed CNN architecture is thoroughly explained. Section \ref{sec:impdetails} explains the implementation details of the experiment. In Section \ref{sec:EvaluationAndComparative}, we elaborate our experiment by describing the evaluations and detailed comparative analysis of the experiment. The proposed scheme is concluded in Section \ref{sec:conclusion}, which includes comprehensive findings and recommendations for further research.


\section{Related work}
\label{sec:rel_work}

In this section, we provide a brief summary of studies in the domain of image-based depth estimation. 

\textbf{Monocular depth estimation}: Alhashim et al. \cite{densedepth} showed that a detailed high-resolution depth maps
how, even for a very simple decoder, our method is able to
achieve detailed high-resolution depth maps. The issue of ambiguous reprojections in depth prediction using stereo-based self-supervision is examined by Watson et al. \cite{depthhints}. They included Depth Hints to counteract such  negative impacts. Laina et al. \cite{FCRN} proposed a robust, single-scale CNN architecture approach that uses residual learning. Ranftl et al. \cite{Midas} recommended the use of principled multi-objective learning to incorporate data from various sources to provide a robust training goal that is invariant to changes in depth range and scale, and emphasize the significance of pretraining encoders on auxiliary tasks. Eigen et al. \cite{MSDN} described a novel technique with the help of two deep network stacks, one produces a rough global prediction based on the entire image and the other makes a more precise local prediction. 

Cantrell et al. \cite{SerialNet} combined the advantages of both semantic segmentation and transfer learning for better depth estimation precision. Hu et al. \cite{SIDE} suggested two changes to the current methods: one deals with the technique of combining features acquired at various scales while the other concerns loss functions, which are employed in training to measure inference errors. Bhat et al. \cite{Adabins2021} suggested an architectural block based on transformers that separates the depth range into bins with adaptively determined centres per image. Yan et al. \cite{CADepth2021} suggested two significant contributions. Firstly, the structure perception module uses the self-attention mechanism to capture long-range dependencies and aggregates discriminative features in channel dimensions. Secondly, the detail emphasis module re-calibrates channel-wise feature maps and selectively emphasises the informative features, resulting in more accurate and sharper depth prediction.

\textbf{Stereo depth estimation}: Duggal et al. \cite{DeepPruner_ROB} demonstrated the case where the cost volume of each pixel is pruned in portions without the need to assess the matching score fully. They constructed a differentiable PatchMatch module to achieve the same. High-resolution imagery processing is a challenge for many state-of-the-art (SOTA) methods due to memory or processing performance limits. Yang et al.\cite{HSM} provided an end-to-end framework that gradually looks for correspondences throughout a coarse-to-fine hierarchy. They accurately predicted disparity for near-range structures with low latency. Komatsu et al. \cite{OctDPS} suggested plane-sweeping strategy created two cost volumes using high and low spatial frequency characteristics while taking the misalignment issue into account. 

Chang et al. \cite{PSMNet} proposed a pyramid stereo matching network consisting of two main modules: the spatial pyramid pooling module for creating cost volume using global context information and the 3D CNN module for regularizing the cost volume. Li et al. \cite{STTR} revisited the issue from a sequence-to-sequence correspondence viewpoint in order to replace the costly pixel-by-pixel building of the cost volume with dense pixel matching that makes use of location information and attention. Anil et al. \cite{SDE} eliminated cost volume construction by learning to match correspondence between pixels with various sets of weights in the dual towers.

\section{Proposed Architecture}
\label{sec:arch}

We propose 2T-UNet, a two-tower UNet with identical contracting paths (encoder), but with different weights and inputs. Fig. \ref{2T-UNet} illustrates the architecture of our 2T-UNet architecture. This network is inspired by the UNet architecture  \cite{UNet}. 2T-UNet differs from standard UNet in three key aspects:

{\textbullet \hspace{1 mm} The standard UNet has a singular contracting (encoder) and expanding (decoder) path, whereas in our 2T-UNet, there exists twin encoders and one decoder.}
    
{\textbullet \hspace{1 mm} We assist stereo-based depth estimation process by providing the network with monocular depth clues of the left view and the stereo image pair.}
    
{\textbullet \hspace{1 mm} The features from the twin encoders are multiplied element-wise to make up the fusion strategy \cite{SDE}. The fusion is performed before concatenating the features from encoders with the upsampled feature map in the decoder.}
    
The contracting paths consist of the repeated application of two $5 \times 5$ convolutions. These convolutions are applied with two padding to preserve the input feature resolution. This is followed by downsampling, which is done by using a $2 \times 2$ max-pool layer with stride equals two. At each downsampling step, we double the channels while reducing the spatial resolution by half. The expanding path is composed of repeated units of $4 \times 4$ up-convolution. Reducing the feature channels to half and doubling the spatial resolution.

A standard UNet is composed of skip connections between the encoder and decoder. These connections serve as a channel for information transfer that helps retain the spatial resolution and object boundaries in the prediction \cite{UNet}. In standard UNet, the encoder feature map is transferred as a whole to the decoder to concatenate with the decoder feature map \cite{UNet}. In our 2T-UNet model, however, there are two encoders and a single decoder. The model architecture necessitates the use of a fusion strategy to combine two encoder features into a single feature map. This fusion is accomplished by multiplying the features of the two encoders element by element before sending them to the decoder for concatenation. Two $5 \times 5$ convolutions with two padding are applied on the concatenated feature maps. Finally, the expansion path outputs a depth map that matches the input’s spatial resolution. In 2T-UNet, the primary encoder receives the left image concatenated with the monocular depth clue as input, while the secondary encoder receives the right image alone. The primary encoder is the only one that is directly connected to the expanding path. The secondary encoder terminates at the last skip connection in the architecture.

\begin{figure*}[t!]
{\includegraphics[width=1in]{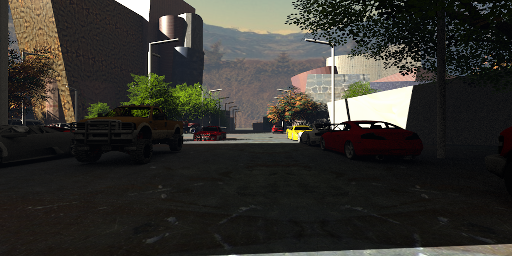}}\hfill
{\includegraphics[width=1in]{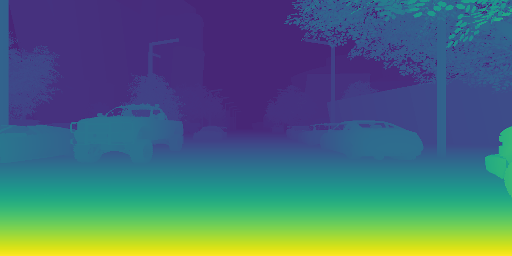}}\hfill
{\includegraphics[width=1in]{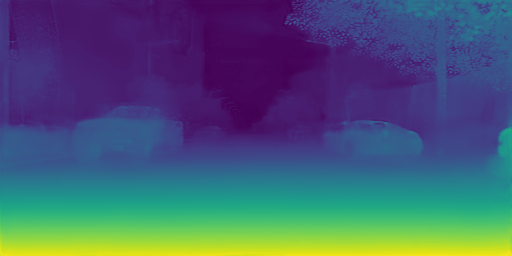}}\hfill
{\includegraphics[width=1in]{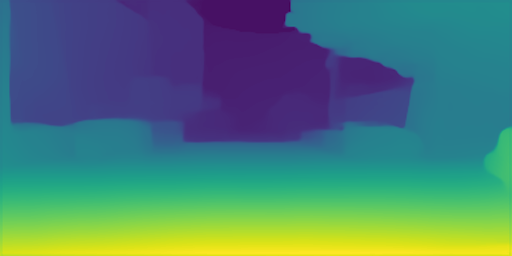}}\hfill
{\includegraphics[width=1in, height = 0.5in]{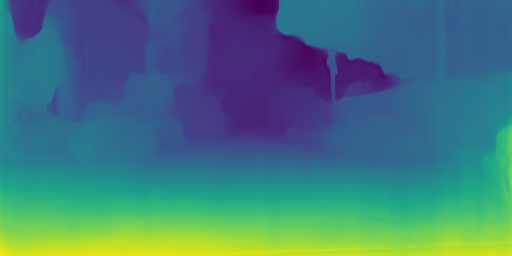}}\hfill
{\includegraphics[width=1in]{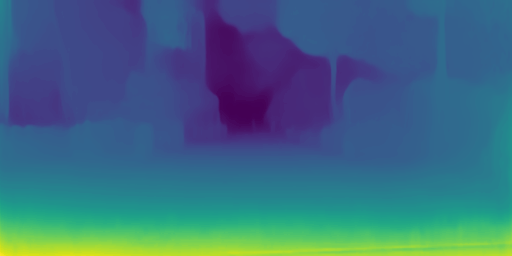}}\hfill
{\includegraphics[width=1in]{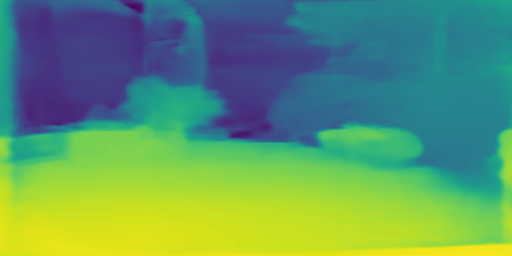}}\hfill

\vspace*{0.5mm}

{\includegraphics[width=1in]{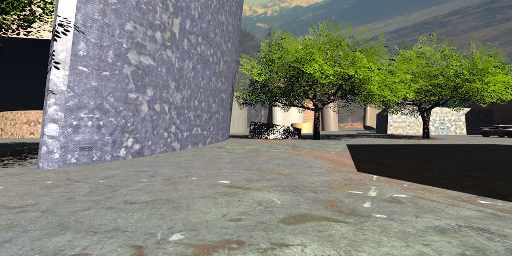}}\hfill
{\includegraphics[width=1in]{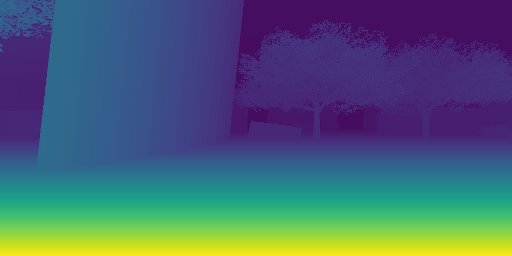}}\hfill
{\includegraphics[width=1in]{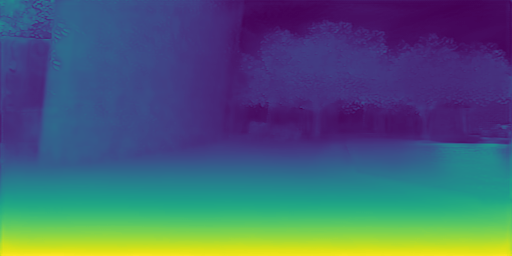}}\hfill
{\includegraphics[width=1in]{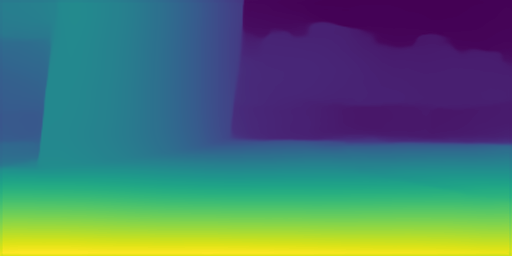}}\hfill
{\includegraphics[width=1in, height = 0.5in]{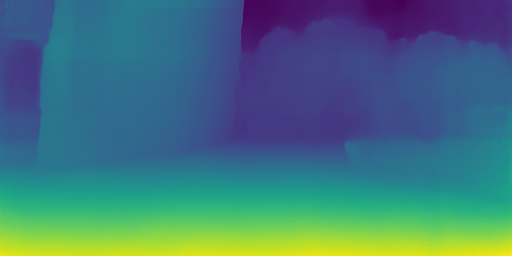}}\hfill
{\includegraphics[width=1in]{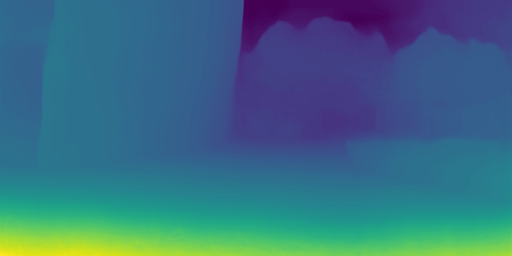}}\hfill
{\includegraphics[width=1in]{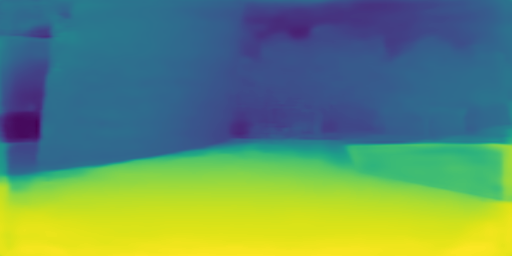}}\hfill

\vspace*{0.5mm}

{\subfloat[Left image]{\includegraphics[width=1in]{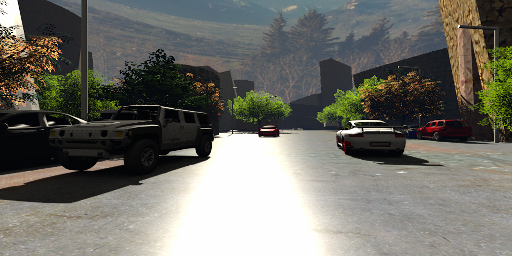}}}\hfill
{\subfloat[Ground truth]{\includegraphics[width=1in]{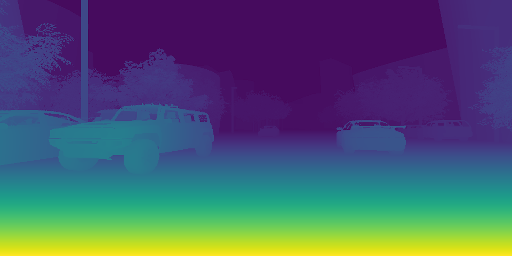}}}\hfill
{\subfloat[2T-UNet (ours)]{\includegraphics[width=1in]{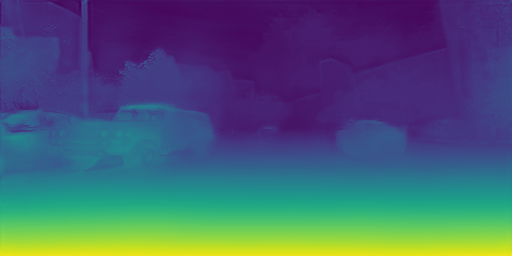}}}\hfill
{\subfloat[MiDaS\cite{Midas}]{\includegraphics[width=1in]{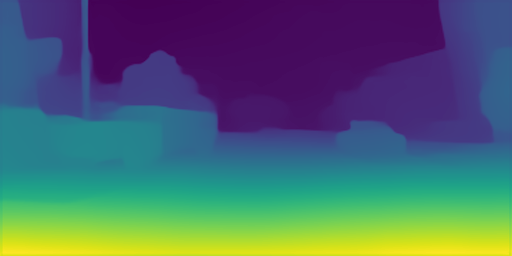}}}\hfill
{\subfloat[CADepth\cite{CADepth2021}]{\includegraphics[width=1in, height = 0.5in]{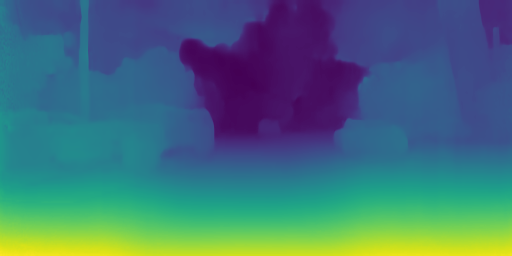}}}\hfill
{\subfloat[Depth Hints\cite{depthhints}]{\includegraphics[width=1in]{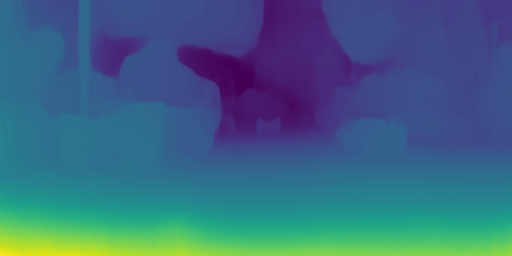}}}\hfill
{\subfloat[AdaBins\cite{Adabins2021}]{\includegraphics[width=1in]{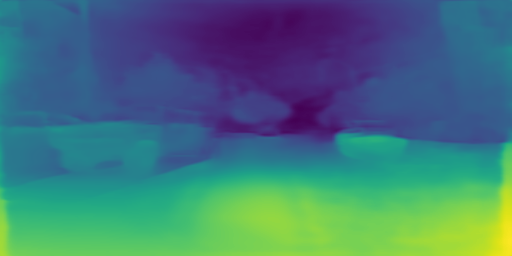}}}\hfill
\caption{Visual comparison results of proposed 2T-UNet method with state-of-the-art monocular depth estimation algorithms.}
\label{visComp}
\end{figure*}

\begin{figure*}[t!]
{\includegraphics[width=1in]{Fig/Fig2&Fig3/LeftView/63.png}}\hfill
{\includegraphics[width=1in]{Fig/Fig2&Fig3/GT/63.png}}\hfill
{\includegraphics[width=1in]{Fig/Fig2&Fig3/2T-UNet/63.png}}\hfill
{\includegraphics[width=1in]{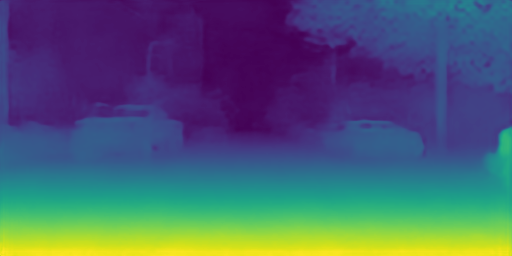}}\hfill
{\includegraphics[width=1in]{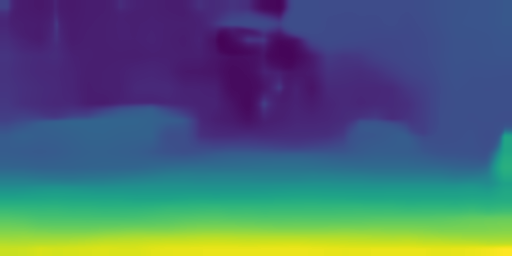}}\hfill
{\includegraphics[width=1in]{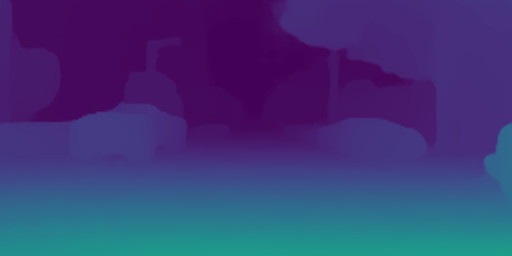}}\hfill
{\includegraphics[width=1in]{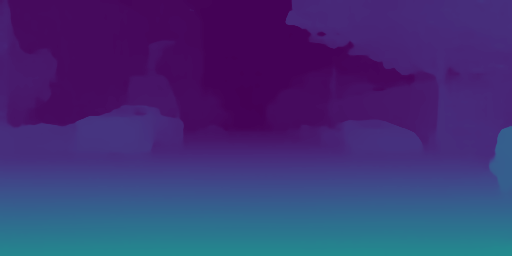}}\hfill

\vspace*{0.5mm}

{\includegraphics[width=1in]{Fig/Fig2&Fig3/LeftView/238.png}}\hfill
{\includegraphics[width=1in]{Fig/Fig2&Fig3/GT/238.png}}\hfill
{\includegraphics[width=1in]{Fig/Fig2&Fig3/2T-UNet/238.png}}\hfill
{\includegraphics[width=1in]{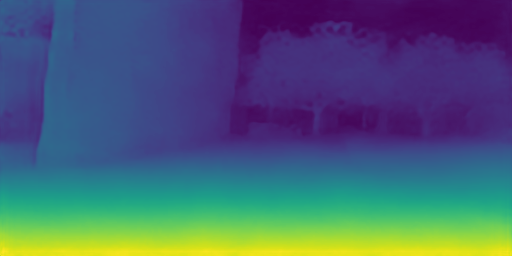}}\hfill
{\includegraphics[width=1in]{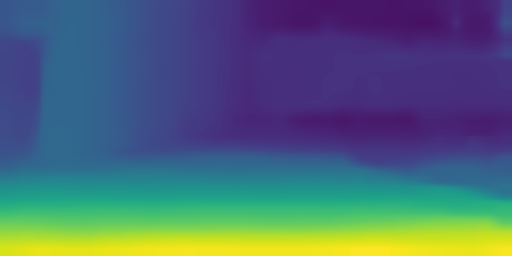}}\hfill
{\includegraphics[width=1in]{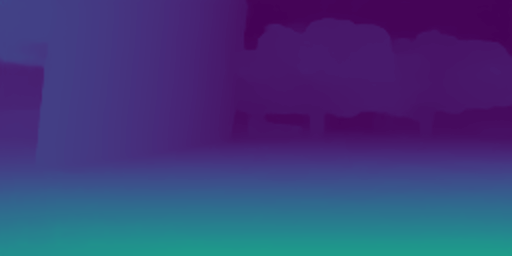}}\hfill
{\includegraphics[width=1in]{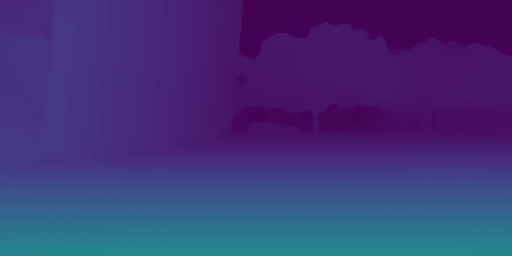}}\hfill

\vspace*{0.5mm}

{\subfloat[Left image]{\includegraphics[width=1in]{Fig/Fig2&Fig3/LeftView/395.png}}}\hfill
{\subfloat[Ground truth]{\includegraphics[width=1in]{Fig/Fig2&Fig3/GT/395.png}}}\hfill
{\subfloat[2T-UNet (ours)]{\includegraphics[width=1in]{Fig/Fig2&Fig3/2T-UNet/395.png}}}\hfill
{\subfloat[SDE-DENet\cite{SDE}]{\includegraphics[width=1in]{Fig/Fig2&Fig3/2T-UNet/395.png}}}\hfill
{\subfloat[HSMNet\cite{HSM}]{\includegraphics[width=1in]{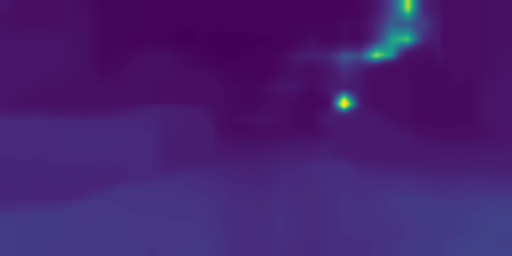}}}\hfill
{\subfloat[DeepROB\cite{DeepPruner_ROB}]{\includegraphics[width=1in]{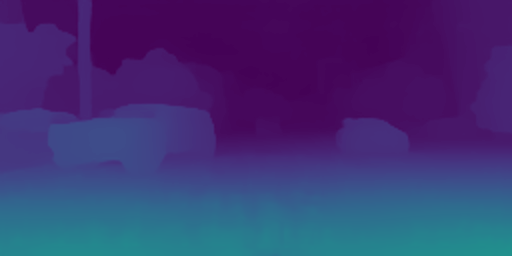}}}\hfill
{\subfloat[PSMNet\cite{PSMNet}]{\includegraphics[width=1in]{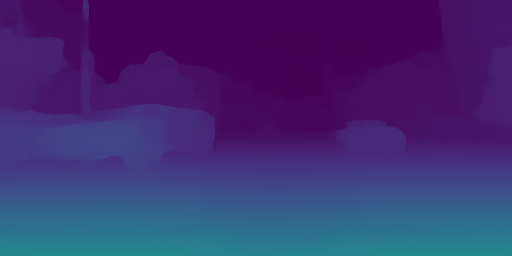}}}\hfill
\caption{Visual comparison results of proposed 2T-UNet method with state-of-the-art stereo based depth estimation algorithms.}
\label{visComp2}
\end{figure*}

\begin{figure*}[t!]
{\includegraphics[width=2.3in]{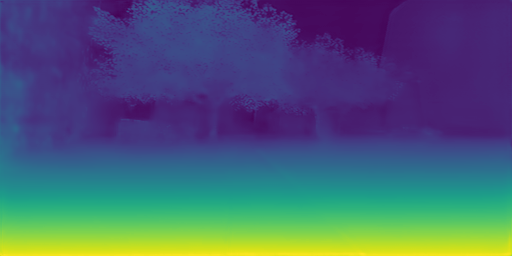}}\hfill
{\includegraphics[width=2.3in]{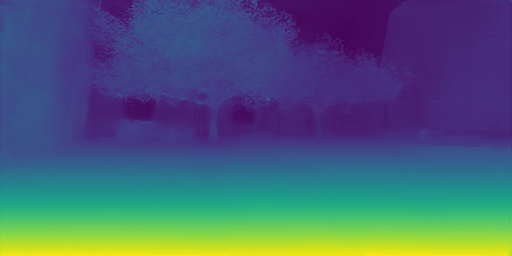}}\hfill
{\includegraphics[width=2.3in]{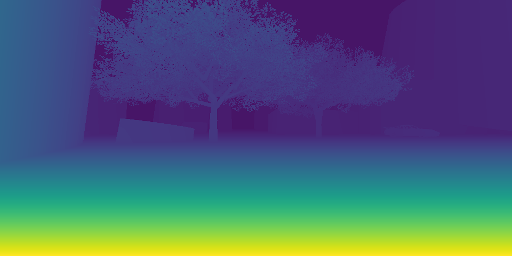}}\hfill
\vspace*{0.5mm}
{\subfloat[Predicted Depth Map without Depth clue]{\includegraphics[width=2.3in]{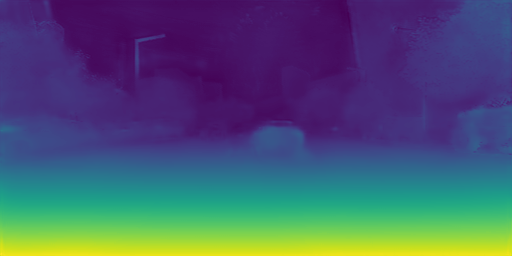}}}\hfill
{\subfloat[Predicted Depth Map with Depth clue]{\includegraphics[width=2.3in]{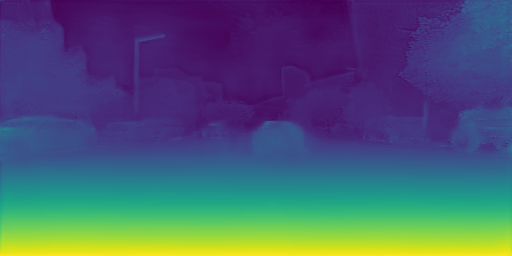}}}\hfill
{\subfloat[Ground Truth]{\includegraphics[width=2.3in]{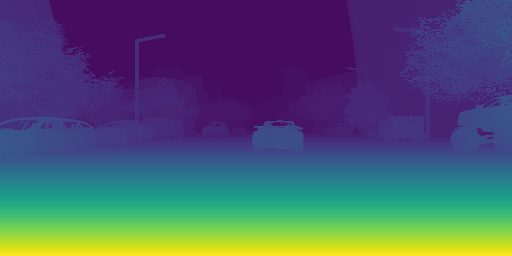}}}\hfill
\vspace*{0.1mm}
\caption{A visual comparison of estimated depth maps using proposed 2T-UNet model considering without depth clue (a) and with depth clue (b). (row1) The depth estimate and feature definitions of building in the foreground is more accurate in (b); (row2) The features of the street lights and smaller vehicles on both sides of (b) are much closer to ground truth (c).\\
}
\label{visComp3}
\end{figure*}

\begin{figure*}[h!]
{\includegraphics[height=0.70in ,width=0.99in]{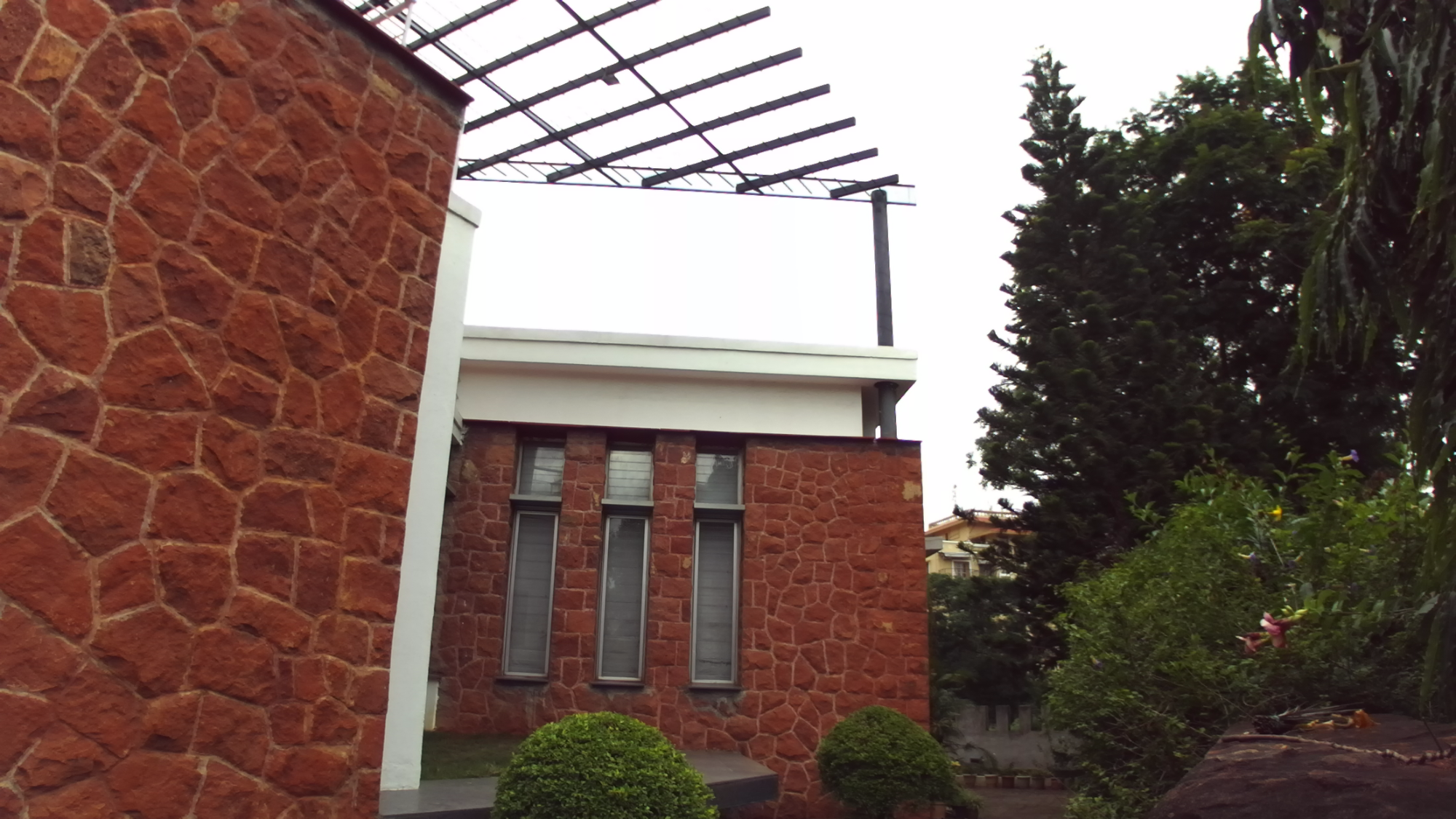}}\hfill
{\includegraphics[height=0.70in ,width=0.99in]{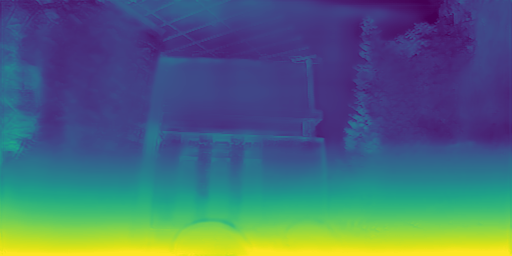}}\hfill
{\includegraphics[height=0.70in
,width=0.99in]{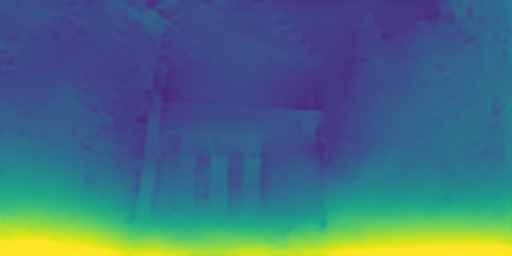}}\hfill
{\includegraphics[height=0.70in ,width=0.99in]{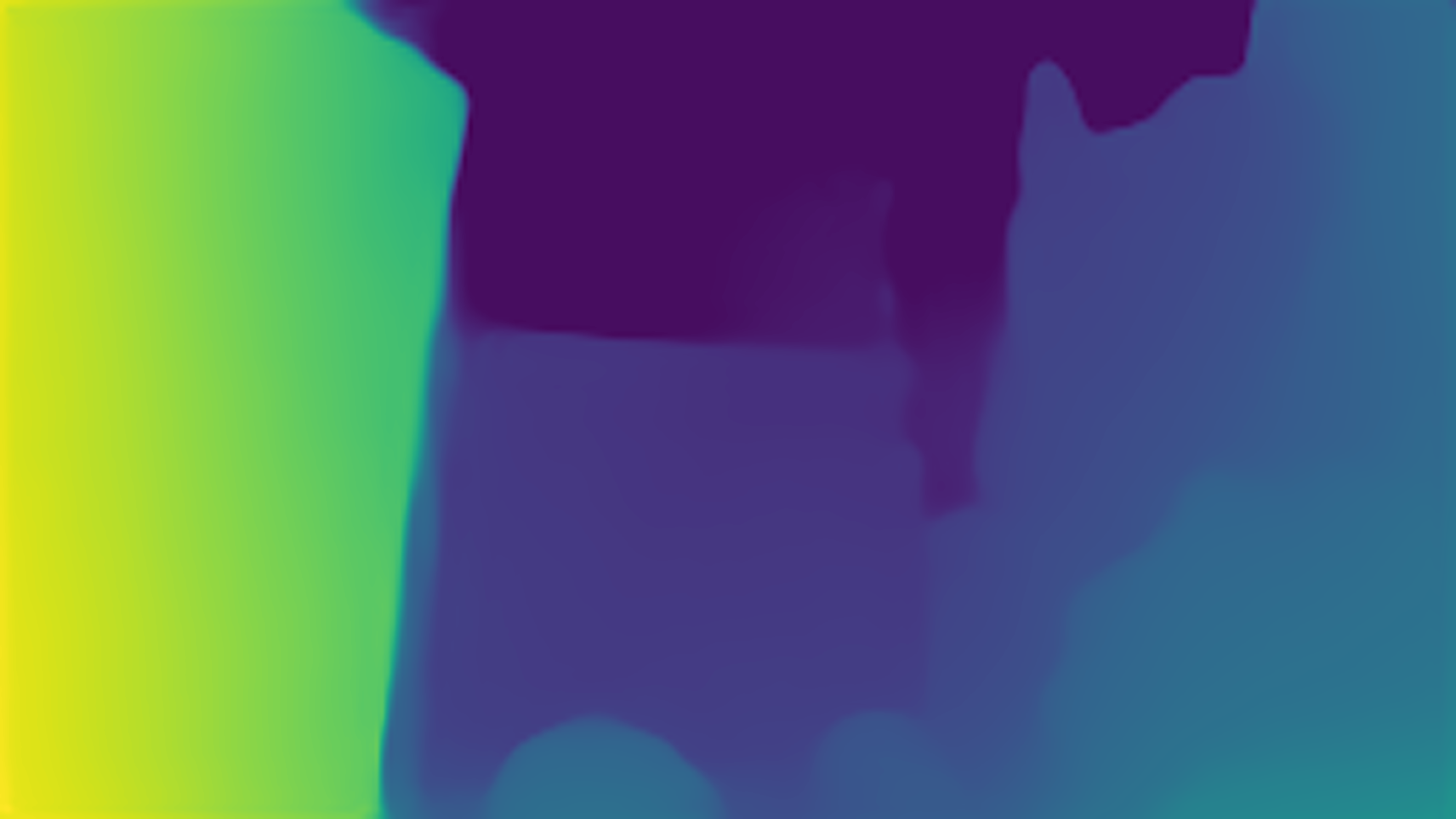}}\hfill
{\includegraphics [height=0.70in
,width=0.99in]{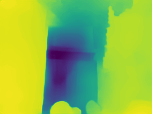}}\hfill
{\includegraphics[height=0.70in ,width=0.99in]{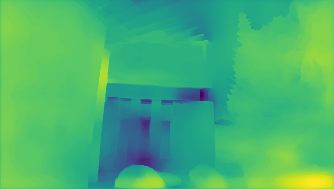}}\hfill
{\includegraphics [height=0.70in
,width=0.99in]{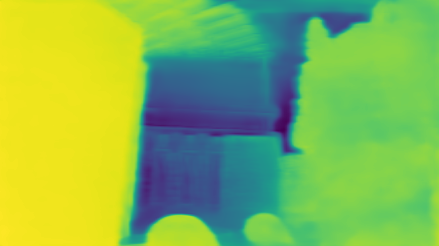}}\hfill

{\includegraphics[height=0.70in ,width=0.99in]{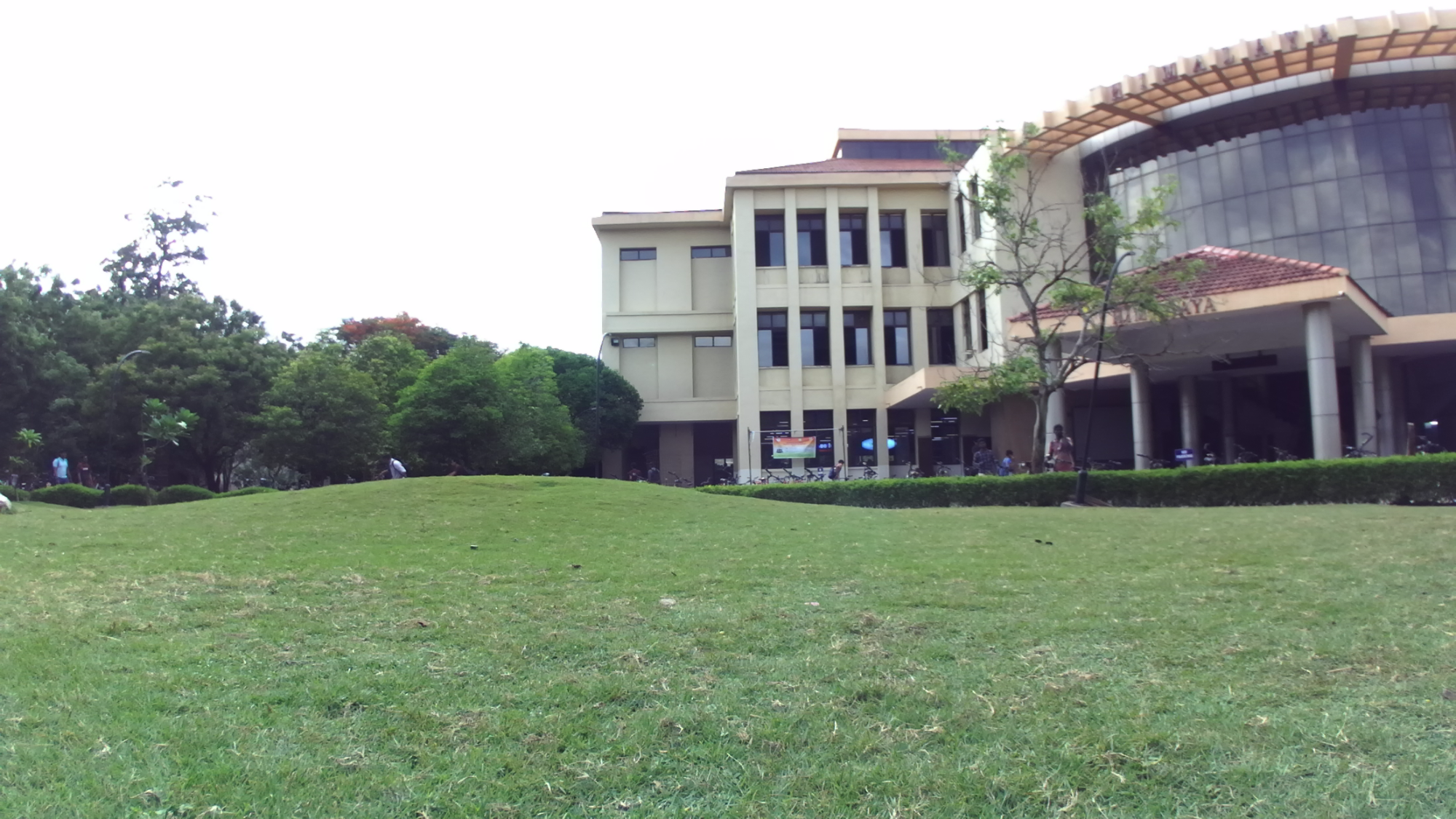}}\hfill
{\includegraphics[height=0.70in ,width=0.99in]{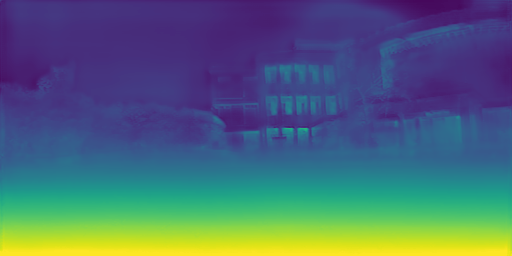}}\hfill
{\includegraphics[height=0.70in
,width=0.99in]{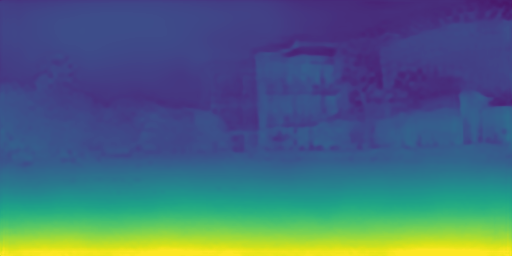}}\hfill
{\includegraphics[height=0.70in ,width=0.99in]{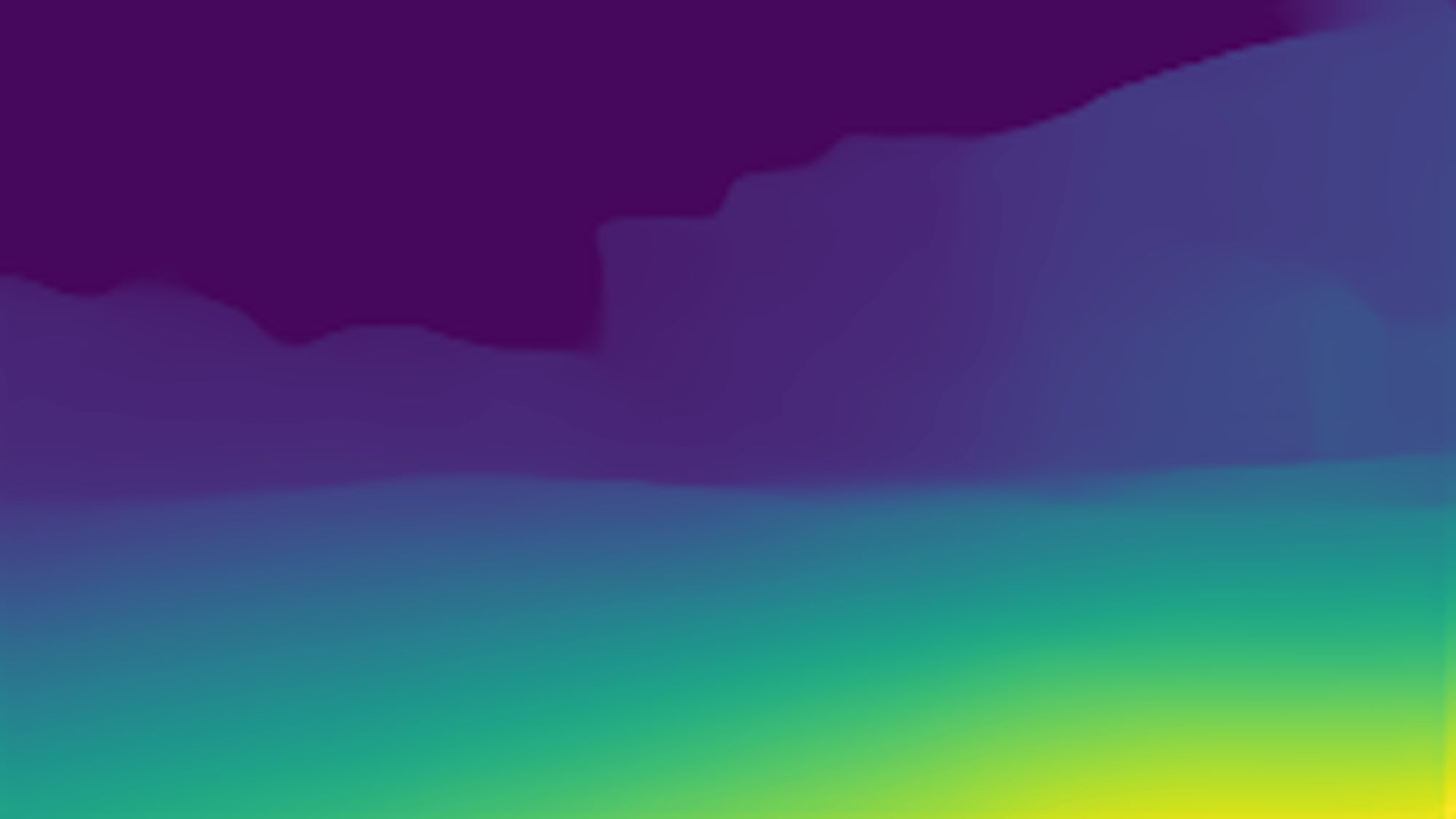}}\hfill
{\includegraphics [height=0.70in
,width=0.99in]{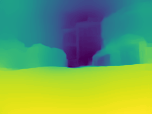}}\hfill
{\includegraphics[height=0.70in ,width=0.99in]{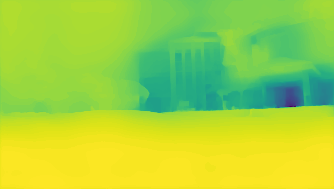}}\hfill
{\includegraphics [height=0.70in
,width=0.99in]{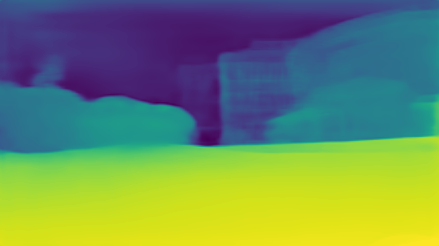}}\hfill

{\subfloat[Left image]{\includegraphics[height=0.70in ,width=0.99in]{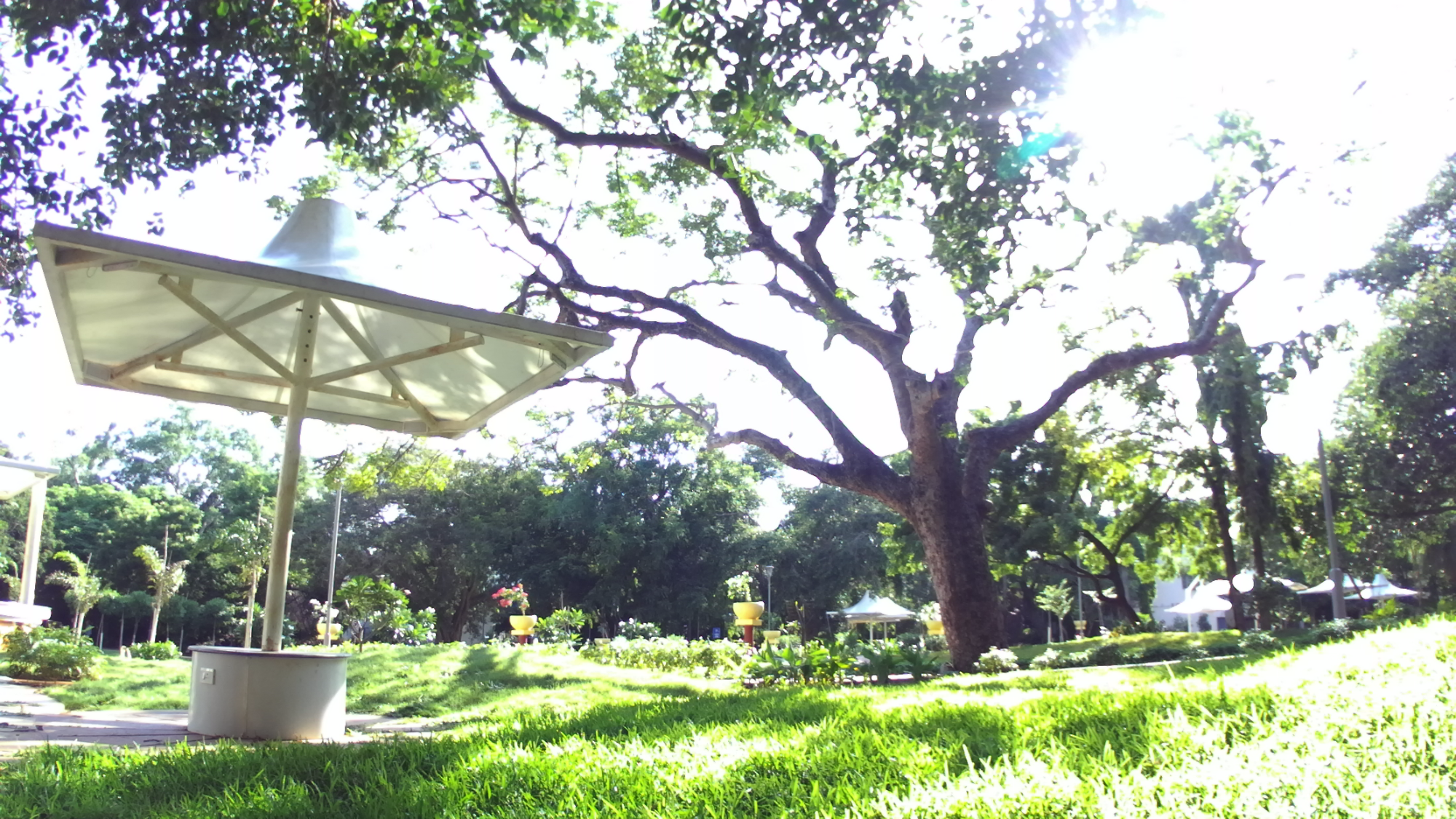}}}\hfill
{\subfloat[2T-UNet(ours)]{\includegraphics[height=0.70in ,width=0.99in]{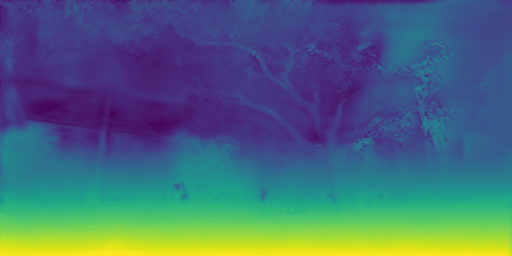}}}\hfill
{\subfloat[SDE-DENet\cite{SDE}]{\includegraphics[height=0.70in ,width=0.99in]{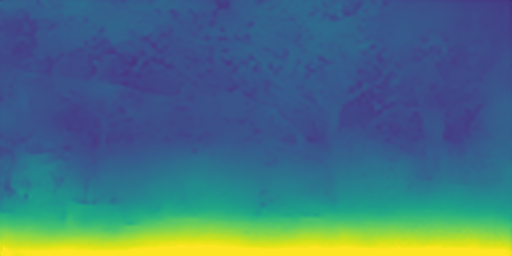}}}\hfill
{\subfloat[MiDaS\cite{Midas}]{\includegraphics[height=0.70in ,width=0.99in]{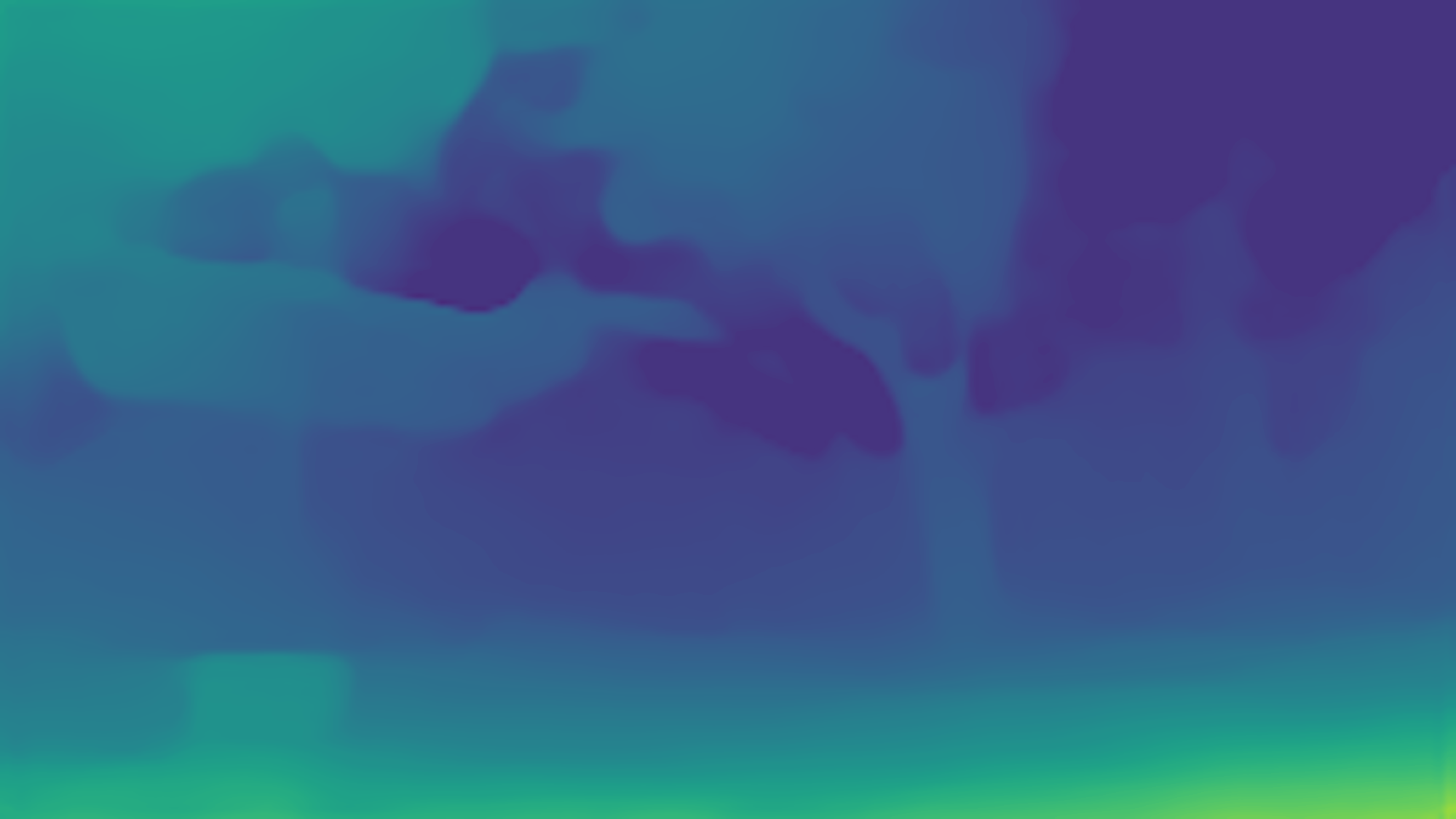}}}\hfill
{\subfloat[SIDE\cite{SIDE}]{\includegraphics[height=0.70in ,width=0.99in]{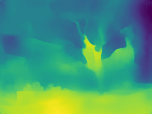}}}\hfill
{\subfloat[DenseDepth\cite{densedepth}]{\includegraphics[height=0.70in ,width=0.99in]{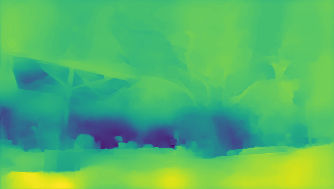}}}\hfill
{\subfloat[AdaBins\cite{Adabins2021}]{\includegraphics[height=0.70in ,width=0.99in]{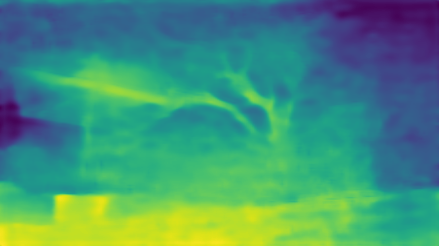}}}\hfill
\caption{Visual comparison results of proposed 2T-UNet
architecture  trained  on Scene flow dataset and tested on natural stereoscopic scenes. We compare with state-of-the-art depth estimation algorithms on $House1$, $Himalaya$, $Canopy3$ scenes \textit{(top to bottom)} from complex natural dataset \cite{Wadaskar2019ME3DHDR}.}
\label{FIG_ZED}
\end{figure*}

\section{Implementation Details}
\label{sec:impdetails}
The 2T-UNet is implemented using PyTorch. The model is trained and evaluated on a high-end Tesla K80 GPU on Google Colaboratory. The training took about 15 hours to complete. Inference time per stereo pair is around 0.22 seconds. We trained our model on synthetic driving dataset from the Scene flow database \cite{SceneFlow}. The dataset is composed of dynamic natural driving scenarios from the first-person perspective of the driver. In addition to cars, the scenes also contain highly detailed objects such as trees, warehouses, and streetlights. The dataset has 4400 pairs of stereo images. The network is trained for 15 epochs using a standard 90:10 train-test split. It is interesting to notice the computational complexity of 2T-UNet model with and without considering depth clues. The total number of trainable parameters remains the same for both these models, while the total parameters increase by 15\% for the model that utilizes depth clues. The inference time of the 2T-UNet model that uses depth clue is 218 ms, while that of the model without depth clue is 217 ms, a difference of approximately 1 ms is noticed. It is safe to say that the computational complexity of the two models is not significantly different.

\section{Evaluation and Comparative Analysis}
\label{sec:EvaluationAndComparative}
The proposed 2T-UNet is compared with monocular and stereo-based depth prediction algorithms. We compare the performance of our architecture with the existing state-of-the-art monocular depth prediction algorithm: Depth Hints \cite{depthhints}, DenseDepth \cite{densedepth}, MiDaS \cite{Midas}, FCRN \cite{FCRN}, SIDE \cite{SIDE}, SerialUNet \cite{SerialNet}, MSDN \cite{MSDN}, AdaBins \cite{Adabins2021}, CADepth \cite{CADepth2021}. In addition, stereo-based depth prediction algorithms are also used for performance comparison: PSMNet \cite{PSMNet}, OctDPSNet \cite{OctDPS}, DeepROB \cite{DeepPruner_ROB}, HSMNet \cite{HSM}, STTR \cite{STTR} and SDE-DENet \cite{SDE}. We use publicly available pre-trained models for evaluating the baseline methods. We separately computed depth maps of right and left images to compare the performance with monocular-based algorithms. Furthermore, a stereo image pair is provided as input for comparison with stereo-based depth prediction methods, with one image serving as a source (reference) and the other as a target. In our model, the stereo image pair is provided as input to the network.

The proposed architecture delivers encouraging results on the Scene flow dataset with high-quality depth maps. In Table \ref{Table-comparison}, quantitative comparison results of 2T-UNet architecture with other baseline methods on the Scene flow dataset are reported. For quantitative comparisons, we use standard error metrics:  absolute relative error ($abs\_rel$), squared relative error ($sq\_rel$), root mean square error (RMSE), average ($log_{10}$) error, threshold accuracy ($\sigma_i$) \cite{densedepth, KenBurns3D}. We also compute mSSIM, which is the mean structural similarity score \cite{3DBLUNet}.

Figures \ref{visComp} and \ref{visComp2} show the visual comparison of the predicted depth maps of the selected networks. We select three scenes from the Scene flow driving synthetic dataset which include tree, car, road, buildings. The scenes include ill-posed areas such as reflective glass, walls and road surfaces. Our proposed algorithm consistently outperforms stereo-based and monocular-based baseline methods, both quantitatively and qualitatively. The 2T-UNet architecture achieves comparable result with SDE-DENet \cite{SDE} method. Our method provides robust depth estimation results particularly in the regions of car windows and wall. The thin structures and object boundaries are clearly retained in our depth estimates. The predicted depth maps with and without the use of depth clues are compared visually in Fig. \ref{visComp3}. The estimated depth map with depth clues has clear depth discontinuity and object structures. The depth structure and fine details are well preserved in scenes with large variations in depth.

In a natural scene with intricate movements, lighting change, and illumination, estimating depth can be difficult. We conduct a visual comparison of our method against existing approaches on challenging nature scenes \cite{Wadaskar2019ME3DHDR}, as shown in Figure 1, to demonstrate its efficacy. For this objective, we employ 2T-UNet architecture that has been trained on the Scene flow dataset. Due to a lack of ground truth data, the quantitative analysis for intricate natural scenes \cite{Wadaskar2019ME3DHDR} is not carried out. 

\section{Conclusion}
\label{sec:conclusion}
In this paper, we explore the issue of stereo depth estimation using a simple CNN and propose an end-to-end Two Tower UNet architecture. The suggested architecture bypasses cost volume construction and, as a result, avoids specifying the disparity range of a scene. Furthermore, the network learns better object boundaries with clarity in depth discontinuity by using depth clues. The proposed 2T-UNet model gives quantitatively and qualitatively better depth predictions at inherently ill-posed regions despite eliminating cost volume construction. We experimentally demonstrated the quality of results using the complex natural scenes and Scene flow dataset. As an extension of the current study, we plan to investigate other feature fusion techniques. Additionally, we would also wish to expand 2T-UNet as a complementary network to additional methods based upon stereo depth estimation. We will further explore the possibilities of extending our work to free-viewpoint rendering and in realistic integration of virtual scenes in augmented reality and 3D environments.

\section{Acknowledgement}
The scientific efforts leading to the results reported in this paper have been carried out under the supervision of Dr. Mansi Sharma, INSPIRE Hosted Faculty, IIT Madras. This work has been supported, in part, by the Department of Science and Technology, Government of India project ‘‘Tools and Processes for Multi-view 3D Display Technologies’’, DST/INSPIRE/04/2017/001853.

\bibliographystyle{plain}
\bibliography{root}

\vspace{-0.3in}
\begin{IEEEbiography}[{\includegraphics[width=1in,height=1.25in,clip,keepaspectratio]{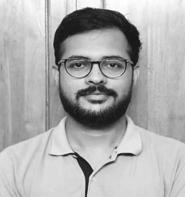}}]{Rohit Choudhary} completed his B.Tech. in Electronics and Communication Engineering, in 2019, from National Institute of Technology Kurukshetra, India. He is currently pursuing M.S. at the Department of Electrical Engineering, Indian Institute of Technology Madras, India. His research interests include 3D Computer Vision, Computational Photography, Deep Learning and 3D Display Technologies. 
\end{IEEEbiography}
\vspace{-0.4in}
\begin{IEEEbiography}[{\includegraphics[width=1in,height=1.25in,clip,keepaspectratio]{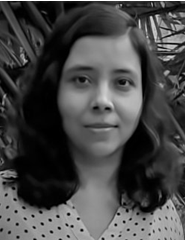}}]{Mansi Sharma} received Ph.D. in Electrical Engineering, in 2017, from IIT Delhi. She received M.Sc. Degree in Mathematics, in 2008, and M.Tech. Degree in Computer Applications, in 2010, from Department of Mathematics, IIT Delhi. She is a recipient of the INSA/DST INSPIRE Faculty award, 2017. Since May 2018, she has been working as an INSPIRE Faculty in the Dept. of Electrical Engineering, IIT Madras. Her research interests include 
Computational Photography, Computational Imaging, Machine Learning, Artificial Intelligence, Virtual and Augmented Reality, 3D Display Technology, Wearable Computing, Visuo-haptic Mixed Reality, Applied Mathematics and Data Science. 
\end{IEEEbiography}

\vspace{-6.3in}

\begin{IEEEbiography}[{\includegraphics[width=1in,height=1.25in,clip,keepaspectratio]{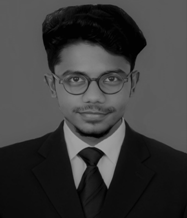}}]{Rithvik Anil} completed his B.Tech. in Mechanical Engineering and M.Tech. in Robotics, in 2021, from India Institute of Technology Madras, India. His research interests include 3D Computer Vision, Autonomous Vehicles and Deep Learning.
\end{IEEEbiography}

\end{document}